\documentclass[10pt,twocolumn,letterpaper]{article}

\usepackage{graphicx}
\usepackage{amsmath}
\usepackage{amssymb}
\usepackage{booktabs}
\usepackage{amsmath}
\usepackage{paralist}
\usepackage{graphics}
\usepackage{color}
\usepackage{float}
\usepackage{multirow}
\usepackage{amssymb}
\usepackage{amsfonts, amssymb} 
\usepackage[ruled,vlined,linesnumbered,resetcount]{algorithm2e}
\usepackage[pagenumbers]{cvpr}

\usepackage[capitalize]{cleveref}
\crefname{section}{Sec.}{Secs.}
\Crefname{section}{Section}{Sections}
\Crefname{table}{Table}{Tables}
\crefname{table}{Tab.}{Tabs.}

\SetKwComment{Comment}{/* }{ */}

\def\cvprPaperID{12558} 
\def\confName{CVPR}
\def\confYear{2022}

\begin{document}

\title{ESTISR: Adapting Efficient Scene Text Image Super-resolution for Real-Scenes}
\author{Minghao Fu$^{1}$, Xin Man$^{1}$, Yihan Xu$^{1}$, Jie Shao$^{1}$\\
$^{1}$University of Electronic Science and
Technology of China, Chengdu, China\\
    {\tt\small \{minghaofu, manxin, yhx\}@std.uestc.edu.cn, \{shaojie\}@uestc.edu.cn,}\\
}

\maketitle

\begin{abstract}

While scene text image super-resolution (STISR) has yielded remarkable improvements in accurately recognizing scene text, prior methodologies have placed excessive emphasis on optimizing performance, rather than paying due attention to efficiency - a crucial factor in ensuring deployment of the STISR-STR pipeline. In this work, we propose a novel Efficient Scene Text Image Super-resolution (ESTISR) Network for resource-limited deployment platform. ESTISR's functionality primarily depends on two critical components: a CNN-based feature extractor and an efficient self-attention mechanism used for decoding low-resolution images. We designed a re-parameterized inverted residual block specifically suited for resource-limited circumstances as the feature extractor. Meanwhile, we proposed a novel self-attention mechanism, softmax shrinking, based on a kernel-based approach. This innovative technique offers linear complexity while also naturally incorporating discriminating low-level features into the self-attention structure. 

Extensive experiments
on TextZoom show that ESTISR retains a high image restoration
quality and improved STR accuracy of
low-resolution images. Furthermore, ESTISR consistently outperforms
current methods in terms of actual running time and
peak memory consumption,
while achieving a better trade-off between performance and efficiency.
\end{abstract}

\section{Introduction}
\label{sec:intro}

Scene text recognition (STR) aims at recognizing text characters in
images, which has revolutionized various downstream tasks such as
license plate recognition \cite{DBLP:conf/eccv/SilvaJ18},
handwriting recognition \cite{DBLP:journals/ijcv/WuYZZL20}, and
scene text visual question answering
\cite{DBLP:conf/icdar/BitenTMGRMJVK19}. However, imperfect
circumstances of image acquisitions always impede the development of
these fields. For example, an image captured under weak light
is hard to be understood in its dark parts, and long focal length
will lead to an undiscriminating image due to its blurry areas.
Therefore, scene text recognition remains a challenge in the wild.

Image super-resolution (SR) is a popular research topic in computer
vision, which aims at recovering low-resolution (LR) images to
high-resolution (HR) correspondence. Since the pioneering work SRCNN
\cite{DBLP:journals/pami/DongLHT16} was proposed, deep learning
based methods have made a breakthrough in the image restoration. Many
studies
\cite{DBLP:conf/eccv/DongLT16,DBLP:conf/cvpr/KimLL16a,DBLP:conf/eccv/ZhangLLWZF18}
have achieved compelling performance in restoration quality, and
recent work
\cite{DBLP:conf/eccv/AhnKS18,DBLP:conf/mm/HuiGYW19,DBLP:conf/cvpr/LiYLYJ019,DBLP:conf/cvpr/TaiY017}
improve the model efficiency to support the wider range of SR
deployments. However, existing SR methods focus on generic image
restoration but not perform well in text images. To address this
issue, TSRN \cite{DBLP:conf/eccv/WangX0WLSB20} first leverages SR as
the pre-processor of STR. SR can retrieve low-resolution (LR) text
images to high-resolution (HR) correspondence, ultimately
alleviates the text recognition difficulty. Experiments show that
SR-processed text images have cleaner and sharper text features than the
original ones. This make them more suitable for neural network recognition.

After that, several methods are proposed to improve recognition
accuracy and super-resolution quality of STISR. TBSRN
\cite{DBLP:conf/cvpr/ChenLX21} highlights the position and content
information in the text-level layout to establish a text-focused
framework. TPGSR \cite{DBLP:journals/corr/abs-2106-15368} introduces the
text priors into STISR to provide categorical guidance for image
recovery. TATT \cite{DBLP:conf/cvpr/MaLZ22} proposes a CNN-based text
attention network to reformulate deformed texts in processed images. Despite the great
advances in SR effectiveness, they use complicated
networks and sometimes require prior knowledge of a large pre-trained model,
which are hard to be applied on resources-limited devices.
Moreover, STISR was essentially born as a pre-processor to strengthen
STR model, but it occupies a large amount of time
from the main task, as shown in Table~\ref{tab:time
comparison}. Results show that each STISR model takes up a lot of
time throughout the inference process, because these models have properties including the large parameter number and the high computation complexity in model forwarding. Hence, it is truly imperative to build an
accurate and efficient STISR model as the qualified plug-in
pre-processor for STR.

\begin{table}[t] 
    \begin{center}
        \begin{tabular}{lcc}
        \specialrule{2pt}{1pt}{1pt}
        \multicolumn{1}{c}{\multirow{2}{*}{Method}} & \multicolumn{2}{c}{CRNN \cite{DBLP:journals/pami/ShiBY17} (6.43ms)}                \\ \cline{2-3}
        \multicolumn{1}{c}{}                        & \multicolumn{1}{c}{SR time (ms)} & SR time occupation \\ \hline
        TSRN \cite{DBLP:conf/eccv/WangX0WLSB20}                                         & \multicolumn{1}{c}{39.41}        & 85.97\%       \\ \hline
        TBSRN \cite{DBLP:conf/cvpr/ChenLX21}                                        & \multicolumn{1}{c}{64.58}        & 90.94\%       \\ \hline
        TPGSR \cite{DBLP:journals/corr/abs-2106-15368}                                        & \multicolumn{1}{c}{59.72}        & 90.28\%       \\ \hline
        TATT \cite{DBLP:conf/cvpr/MaLZ22}                                         & \multicolumn{1}{c}{68.49}        & 91.42\%       \\ \specialrule{2pt}{1pt}{1pt}
        \end{tabular}
    \end{center}
    \caption{Comparison of average running time in different STISR networks. We
choose the classic text recognition model CRNN
\cite{DBLP:journals/pami/ShiBY17} as the subsequent recognizer.
SR time is calculated from STISR model inference on TextZoom
\cite{DBLP:conf/eccv/WangX0WLSB20}. SR time occupation denotes the percentage of SR
processing time during the entire recognition.}
    \label{tab:time comparison}
\end{table}

A handicap of current STISR methods is the lack of efficient and
powerful feature extractors. Most representative single image
super-resolution (SISR) models essentially are composed of variants
of convolutional layers. Meanwhile, STISR methods tend to adopt CNN-based modules as the feature extractors before sequence-to-sequence models. However, traditional CNN-based modules are very complex and inefficient for processing images with large channel. They do not consider model simplicity when interacting locally across channels and spatial dimensions. To make the process of providing feature maps into a sequential model more lightweight, we propose a re-parameterized
inverted residual block (RIRB) as our feature extractor backbone.
Due to the preservation of the manifold of interest and the benefits of over-parameterization, RIRB exhibits a powerful feature extracting capacity. During the inference stage, RIRB maintains running speed as fast as $3\times3$
convolution by the structural re-parameterization strategy
\cite{DBLP:conf/cvpr/Ding0MHD021}.

After reviewing the architecture of all the current STISR methods, we discovered that Transformer-based modules play a significant role as components of
\cite{DBLP:conf/cvpr/ChenLX21,DBLP:conf/cvpr/MaLZ22,DBLP:journals/corr/abs-2106-15368}.
Transformer has achieved great success in processing sequential
information and capturing cross-domain dependencies, but it is
extremely computationally expensive for long input sequences as a
result of quadratic space and time complexity in its multi-head
self-attention. Based on the kernel-based method, we propose a novel
self-attention structure with linear complexity. Besides,
in order to make efficient Transformer adapt to STISR, we introduce
low-level information into self-attention matrix, as the third-party
similarity factor between $Q$ and $K$. We utilize a CNN-based module with a lightweight architecture to produce a matrix that introduces low-level prior knowledge into the similarity computation of scale dot product. This low-level feature generator offers a robust prior knowledge to modify locality discrimination and feature interactions in spatially embedded sequences.

Our main contributions can be summarized as follows:
\begin{compactitem}
    \item A re-parameterized inverted residual block (RIRB) is proposed to effectively extract significant feature to be fed into sequence model. Moreover,
RIRB keeps a practically fast running speed as same as $3\times3$
convolution.
    \item Our proposal involves an innovative and efficient self-attention, termed as softmax shrinking, which refomulates the scale dot product to reduce the complexity of self-attention from quadratic to linear.
    \item Building upon this, we have developed an efficient scene text image super-resolution (ESTISR) network that achieves a superior trade-off between performance and efficiency. Compared with existing methods, ESTISR reduces the running time by up to 60\%.
\end{compactitem}

\begin{figure*}[t]
\centering
\includegraphics[width=0.8\textwidth]{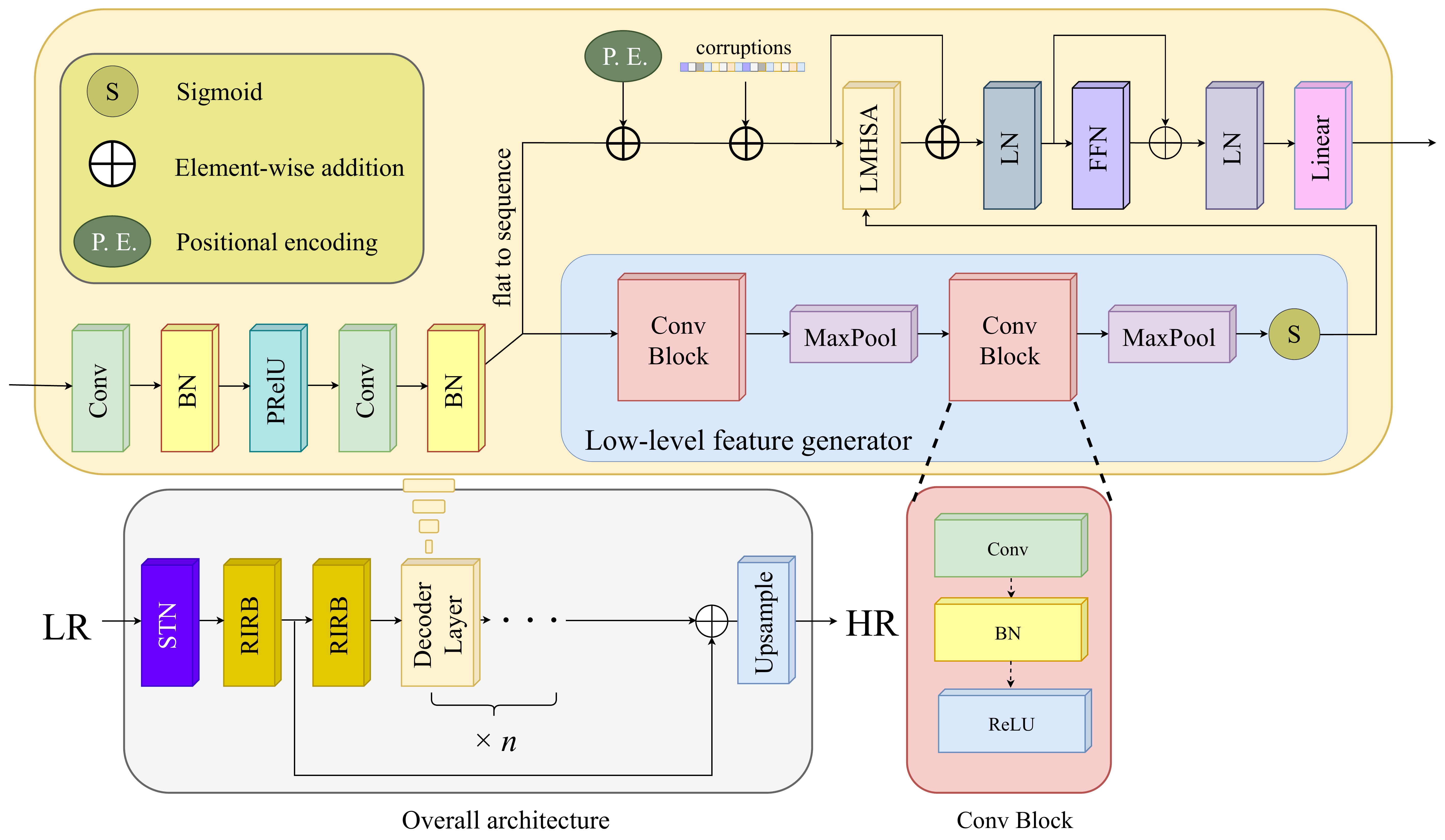}
\caption{Overall network architecture of the proposed efficient
scene text image super-resolution (ESTISR). LMHSA represents multi-head self-attention with linear complexity, LN denotes layer normalization, BN denotes batch normalization, and FFN is the feed-forward network \cite{DBLP:conf/nips/VaswaniSPUJGKP17}.} \label{fig:network}
\end{figure*}

\section{Related Work}

In this section, we provide an overview of how previous SR models
achieved excellent performance and improve efficiency. Then, we
go down to the most relevant works that seek to STISR addressing the
STR accuracy degradation in low-resolution images. Finally, we
introduce recent efficient Transformers including sparse methods and
kernel methods.

\textbf{Single image super-resolution.} In recent years, deep
learning based methods for single image super-resolution (SISR)
\cite{DBLP:journals/pami/DongLHT16,DBLP:conf/cvpr/KimLL16a,DBLP:conf/cvpr/LimSKNL17,DBLP:conf/eccv/ZhangLLWZF18,DBLP:conf/cvpr/LedigTHCCAATTWS17}
have been characterized by a variety of advancements since the
pioneering work of SRCNN \cite{DBLP:journals/pami/DongLHT16}. By
introducing residual learning, attention mechanism, and CNNs into
low-level vision tasks, numerous SR methods have made progresses in
image quality measured by PSNR/SSIM. However, due to the dilemma of
huge memory costs and limited computing resources on mobile devices,
it is imperative to develop efficient SR
\cite{DBLP:conf/eccv/AhnKS18,DBLP:conf/mm/HuiGYW19,DBLP:conf/eccv/DongLT16,DBLP:conf/eccv/LiuTW20,DBLP:conf/cvpr/KongLLLHBCF22}
for real-world usages. Previous efficient SISR models tend to reduce
FLOPs until Zhang \etal \cite{DBLP:conf/mm/ZhangZZ21} point out that
FLOPs may not be an equivalent judgment of running speed. They
propose a straightforward convolution network with structural
re-parameterization \cite{DBLP:conf/cvpr/Ding0MHD021} and achieve
the fastest inference speed in mobile devices. Nevertheless, existing SISR methods do not perform satisfactorily on scene text images.

\textbf{Scene text image super-resolution.} There has been great
progress in the field of scene text recognition (STR) in recent
years. CRNN \cite{DBLP:journals/pami/ShiBY17} first adopts recurrent
neural network to capture scene text semantic and deduce
characteristic substance. After that, ASTER
\cite{DBLP:journals/pami/ShiYWLYB19} and MORAN
\cite{DBLP:journals/pr/LuoJS19} introduce attention mechanism to
provide rectification cross text content. However, scene text
recognition in low-resolution (LR) images remains a challenging task
due to their blurry texture and indistinct contours. Hence, Wang
\etal \cite{DBLP:conf/eccv/WangX0WLSB20} build TextZoom, a
text-focused SR dataset which is collected by different focal
lengths, which aim at using image super-resolution
as the pre-processor before the STR process to alleviate the recognition
difficulty. TBSRN \cite{DBLP:conf/cvpr/ChenLX21}
concentrates on character region and text identification by
position-aware module and context-aware module. TPGSR
\cite{DBLP:journals/corr/abs-2106-15368} injects text priors into the
SR module to provide categorical guidance, and furthermore, TATT
\cite{DBLP:conf/cvpr/MaLZ22} proposes a CNN-based text attention to
address the variation caused by text deformations. Nevertheless, the
methods above consume large amounts of computational resource and
memory, which is unfriendly for deploying practical applications.
Therefore, our ESTISR aims to modify STISR from an efficiency
perspective, and retains its effectiveness on both recognition
accuracy and image quality.

\textbf{Efficient self-attention.} After years of development, the
Transformer architecture \cite{DBLP:conf/nips/VaswaniSPUJGKP17} has
been successfully applied to computer vision tasks
\cite{DBLP:conf/iclr/DosovitskiyB0WZ21,DBLP:conf/eccv/CarionMSUKZ20,DBLP:conf/iccv/LiuL00W0LG21}.
For the power of capturing contextual information in fore-and-aft
character sequence, all recent STISR
\cite{DBLP:conf/cvpr/ChenLX21,DBLP:journals/corr/abs-2106-15368,DBLP:conf/cvpr/MaLZ22} utilize multi-head self-attention as fundamental components to improve learning
global texture. However, quadratic space and time complexities
of scale dot product impede their practical applications in STR. An
intuitive solution is to modify self-attention as the new
architecture backbone. Recent methods can be broadly divided into
two categories, sparse attention and linear attention. Sparse
attention
\cite{DBLP:conf/nips/ZaheerGDAAOPRWY20,DBLP:conf/icml/TayBYMJ20,DBLP:conf/iclr/KitaevKL20}
aim to reformulate the attention matrix, and kernel-based methods
\cite{DBLP:conf/iclr/ChoromanskiLDSG21,DBLP:conf/iclr/QinSDLWLYKZ22,DBLP:conf/iclr/Peng0Y0SK21}
have pushed the acceleration of Transformer in various fields.
However, an in-depth examination about the literature shows that
they either lack theoretical validity or are ineffective empirically
in low-level vision. Our work addresses this issue by incorporating low-level knowledge into our proposed efficient self-attention mechanism.

\section{Methodology}

\subsection{Network Architecture}

The architecture of the proposed ESTISR is depicted in
Figure~\ref{fig:network}. ESTISR applies a sequential backbone to
minimize memory consumption. We first apply a spatial Transformer network (STN)
\cite{DBLP:conf/nips/JaderbergSZK15} on low-resolution (LR) images
to tackle the misalignment problem. After understanding input images
through 2 RIRBs as feature extractors, cascaded decoder layers process
flattened feature maps for capturing global self-attention. We
perform an upsampling operation at the end via the convolution layer and
pixel-shuffle layer, ultimately getting $\times2$ output images.

\subsection{Re-parameterized Inverted Residual Block}

\begin{figure*}[t] \scriptsize
    \centering
    \includegraphics[width=0.8\textwidth]{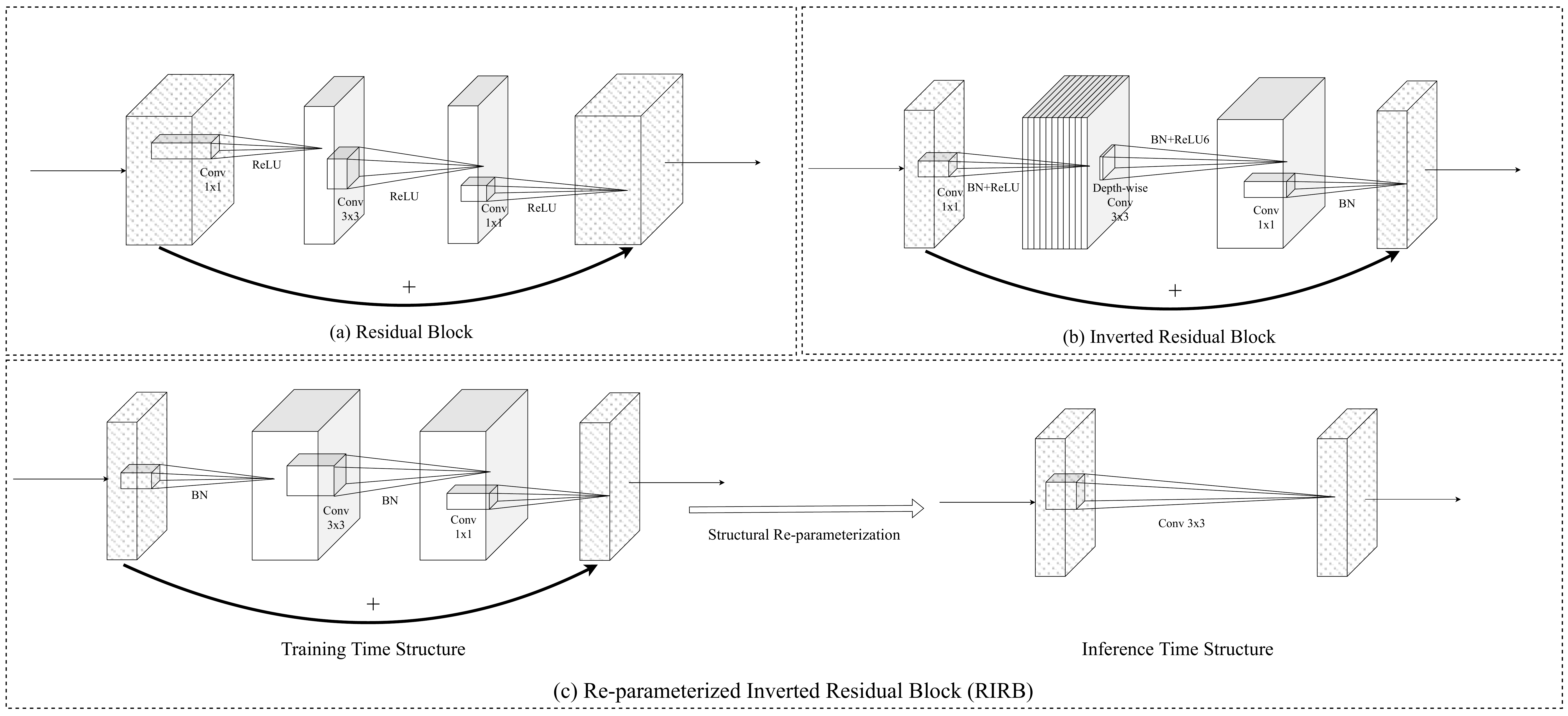}
    \caption{Residual block (a), inverted residual block (b) and our
re-parameterized inverted residual block (c).}
    \label{fig:RIRB}
\end{figure*}

\subsubsection{Potential of Sequential Linear Layer Merging}

Structural re-parameterization is a prevalent methodology in the domain of Super Resolution (SR), serving as a lightweight strategy for employing a CNN backbone. Nevertheless, existing approaches \cite{DBLP:conf/mm/WangDS22, DBLP:conf/mm/ZhangZZ21} primarily focus on multi-branch merging while neglecting the sequential linear layers merging aspect. In our investigation, we have identified that each structural re-parameterization technique proposed in the works of \cite{DBLP:conf/mm/WangDS22, DBLP:conf/mm/ZhangZZ21} possesses an essential property of invertibility. This key characteristic enables the facilitation of sequential linear layers merging, thereby augmenting the potential of the SR framework.

Derived from the original residual block proposed by He et al. \cite{DBLP:conf/cvpr/HeZRS16} as depicted in Figure~\ref{fig:RIRB}(a), the inverted residual block introduced by Sandler et al. \cite{DBLP:conf/cvpr/SandlerHZZC18} in Figure~\ref{fig:RIRB}(b) has demonstrated remarkable efficacy in achieving model lightweightness. However, it does come with inherent limitations that constrain the model's representational capacity. To strike a balance between the desired model complexity and runtime efficiency, we propose a novel re-parameterized inverted residual block (RIRB) that capitalizes on the intrinsic associativity and linearity of sequential linear layers, with removing activation in inverted residual, as illustrated in Figure~\ref{fig:RIRB}(c). Despite the inherent limitation posed by the absence of non-linear activation, our innovative re-parameterization approach facilitates a reduction in network depth through the fusion of sequential linear layers. This strategic consolidation allows us to take advantage of the computational efficiency demonstrated by standard $3\times3$ convolutions, particularly within the context of mobile devices \cite{DBLP:conf/mm/ZhangZZ21}. Notably, to mitigate the information loss caused by the activation reduction, we decrease the expanding ratio in the bottleneck to 2, in contrast to the original ratio of 6.

 Our RIRB differs from existing methods such as the RepSR block and edge-oriented convolution block (ECB) \cite{DBLP:conf/mm/WangDS22, DBLP:conf/mm/ZhangZZ21} in several ways. While these techniques focus on lightening multi-branch structures, they do not fully exploit the potential of sequential linear layers. In contrast, RIRB leverages the flexibility of channel transformation to preserve the desired manifold with a simple structure. Moreover, compared to image classification techniques like ACNet \cite{DBLP:conf/iccv/DingGDH19} and RepVGG \cite{DBLP:conf/cvpr/Ding0MHD021}, which are designed for high-level vision, RIRB is more adaptive to low-level vision tasks.

\subsubsection{Structural Re-parameterization Details}

We will illustrate the underlying principles behind the indispensability of sequential linear layers merging in order to unleash the potential of structural re-parameterization. By training a more intricate formulation, we are able to enhance the representational capacity of the standard convolution. 

We now describe how to re-parameterize RIRB into a
standard $3\times3$ convolution. After re-parameterization, output
feature $F$ can be calculated by the final weight and bias
$\{W_{rep}, B_{rep}\}$ of convolution:
\begin{align}
    F &= W_{rep} \circledast X + B_{rep}.
\end{align}
Below we show the calculation of $\{W_{rep}, B_{rep}\}$:
\begin{align}
    W_{0,1} &= perm(W_0) \circledast W_1, \\
    B_{0,1} &= (W_1) \circledast (B_0 \ pad \ B_0) + B_1,
\end{align}
where $\{W_0, B_0\}$ and $\{W_1, B_1\}$ represent weight and bias of
first Conv$1\times1$ and second Conv$3\times3$.
$\{W_{0,1},B_{0,1}\}$ denotes the weight and bias of their
combination after re-parameterization. $perm$ means exchanging the
first and second dimensions of the convolution kernel. Then, the
problem is converted to a Conv$3\times3$-Conv$1\times1$ combination.
\begin{align}
    W_{rep} = res(bmm(res(W_{0,1}), res(W_2))) + W_I, \\
    B_{rep} = mm(res(W_{0,1}), B_{0,1}) + B_2 + B_I,
\end{align}
where $\{W_2, B_2\}$ and $\{W_I, B_I\}$ represent the third
Conv$3\times3$ and the identity. $res$ represents reshaping kernel,
and $mm (bmm)$ is (batch) matrix multiplication. The side-way
identity connection (channel-wise operation) is supposed to conduct
a direct addition. However, the method how to re-parameterize it has
not been developed. We regard the identity connection as a standard
$3\times3$ convolution with sparse kernel. In this way, RIRB is re-parameterized
into a standard Conv$3\times3$. Although RIRB is more expensive for
training than the standard Conv$3\times3$, an efficient inference is
our drive to improve the model performance. The whole pseudo code
can be viewed in appendix.


\subsection{Self-attention with linear complexity}

In this section, we empirically analyze the choice of patch embedding and spatial embedding in image Transformer. We then introduce softmax shrinking, a novel linear self-attention mechanism that incorporates a low-level feature weight matrix. We provide detailed explanations of each component to enhance comprehension of their respective roles and functionalities.

\subsubsection{Patch-wise or Spatial-wise Embedding}

\begin{figure}[t]
\centering
\includegraphics[width=0.4\textwidth]{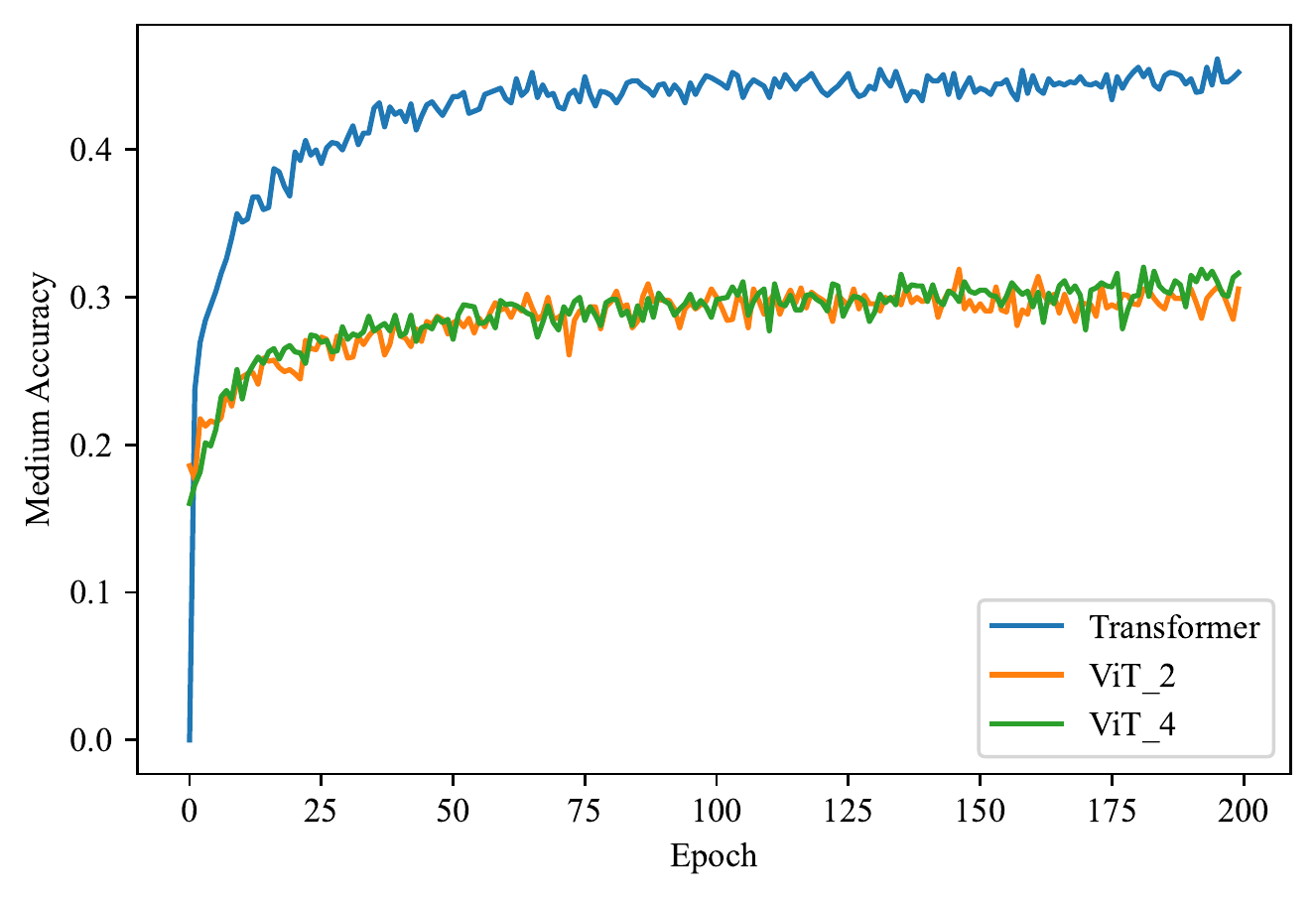}
\caption{Accuracy comparison of patch embedding and spatial
embedding.} \label{fig:patch_spatial}
\end{figure}

We first explore the right way of embedding images in Transformer.
After TSRN \cite{DBLP:conf/eccv/WangX0WLSB20} emerges, recent STISR
methods adopt the spatial-wise embedding for the later dot-product
self-attention calculation. This means each embedded token represents the channels at one position. Patch-wise embedding is a method of representing an image as a sequence of patches like the vision transformer \cite{DBLP:conf/iclr/DosovitskiyB0WZ21}, which has shown good performance in various high-level vision tasks. We select TBSRN
\cite{DBLP:conf/cvpr/ChenLX21}, a Transformer based super-resolution
network as the baseline. As shown in Figure~\ref{fig:patch_spatial},
``ViT" denotes using patch embedding before the query, key and value
fed into self-attention, and the suffix number represents the patch
size. The other configurations follow the original TBSRN.

Regarding Figure~\ref{fig:patch_spatial}, our analysis revealed that using patch-wise embedding in ViT has a detrimental effect on the STISR task. Spatial-wise embedding appears to be better suited for extracting low-level information, due to its ability to adapt to small dimensionality. In contrast, patch-wise embedding tends to focus more on long-range dependencies. Furthermore, the Transformer model heavily relies on local information, and if the sequence length is reduced, the positional embedding's impact is significantly diminished, which affects the Transformer's effectiveness on the sequence. In summary, our results indicate that spatial-wise embedding outperforms patch-wise embedding, and that low-level information is more critical than high-level information for achieving effective STISR.

\subsubsection{Linear Self-attention}

The vanilla multi-head self-attention \cite{DBLP:conf/nips/VaswaniSPUJGKP17} has a quadratic scaling with sequence length, resulting in a large computation burden when $n$ is large. To address this issue and improve the performance of ViT on STISR tasks, we propose a novel linear self-attention method, named softmax shrinking that reduces complexity from $O(n^{2})$ to $O(n)$. Specifically, we introduce a low-level feature generator as an external factor in scale dot product. Transformer-based modules and improves the efficiency of the model.

\begin{figure*}[t]
\centering
\includegraphics[width=1.0\textwidth]{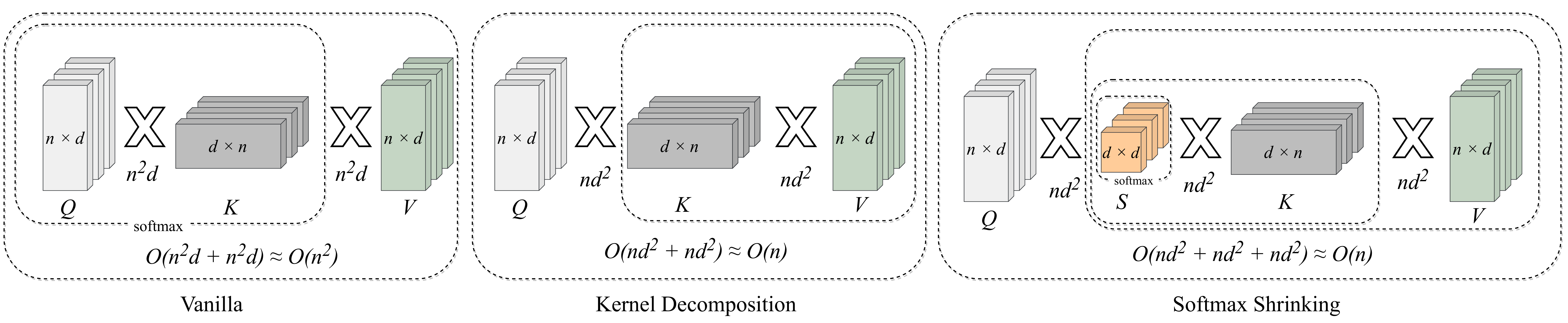}
\caption{Illustration of the computations for vanilla self-attention
with quadratic complexity (left), linear self-attention via kernel
decomposition (middle), and our linear self-attention via softmax
shrinking (right).} \label{fig:linear attention}
\end{figure*}

\paragraph{Kernel-based method}

First we introduce the kernel-based method in previous work. Given queries $Q$, keys $K$ and values $V \in \mathbb{R}^{n \times
d}$, the general form of self-attention is
\begin{equation}
    Attention(Q, K, V) = softmax(\frac{QK^T}{\sqrt{d}})V.
    \label{eq:vanilla}
\end{equation}
The vanilla Transformer uses linear projection kernel to extract
latent feature. We also apply a linear transformation to process
incoming data. In Figure~\ref{fig:linear attention}, we could easily
get that the attention map is an $n \times n$ matrix and the
computation complexity is $O(n^2)$. Various linear self-attention
mechanisms
\cite{DBLP:conf/iclr/ChoromanskiLDSG21,DBLP:conf/iclr/QinSDLWLYKZ22,DBLP:conf/cvpr/ChenLX21}
attempt to tackle the quadratic complexity of such operation by a
kernel decomposition. Katharopoulos \etal
\cite{DBLP:conf/icml/KatharopoulosV020} interpret $softmax(\cdot)$
as a general similarity matrix between $Q$ and $K$. Thus, we
reconstitute self-attention as a similarity matrix of $Q$ and $K$ as
follows:
\begin{equation}
    Attention(Q, K, V) = similar(Q, K)V.
    \label{eq:similar}
\end{equation}
Then, similarity matrix can be described as the multiplication of
representations of $Q$ and $K$ after feature extraction:
\begin{equation}
    Attention(Q, K, V;\phi) = (\phi(Q) \phi(K)^T)V,
    \label{eq:decompose}
\end{equation}
where $\phi(\cdot)$ denotes the latent representation of input data.
After that, dot-product computation order can be rearranged due to
associativity of matrix multiplication:
\begin{equation}
    Attention(Q, K, V;\phi) = \phi(Q)(\phi(K)^TV).
    \label{eq:rearrange}
\end{equation}
As visually illustrated in Figure~\ref{fig:linear attention}. instead of explicitly computing the attention matrix $A = QK^T \in
\mathbb{R}^{n \times n}$, in Eq.~\eqref{eq:rearrange} we first
calculate $\phi(K)^T V \in R^{d \times d}$, and then multiply
$\phi(Q) \in R^{N \times d}$. By using this trick, we only incur a
computation complexity of $O(nd^2)$. Note that, in spatial-wise
embedded images, the feature dimension of one head $d$ in sequence
is always much smaller than the input sequence length $n$, so we can
safely omit $d$ and achieve computation complexity of $O(n)$, as
visually illustrated in Figure~\ref{fig:linear attention}.

\paragraph{Softmax shrinking}

Although kernel-based methods largely decrease computation of
self-attention, such lightweight schemes are originally designed and
applied for natural language processing, and not performs
satisfactorily in the STISR task that involves sequential text and
low-level feature information. Hence, we propose a novel
$softmax(\cdot)$ shrinking based on kernel-based method, to add
low-level feature information into self-attention for capturing more
localized information.

In \cite{DBLP:conf/iclr/QinSDLWLYKZ22}, the $softmax(\cdot)$
operator is considered as the key property to maintain approximation
capacity of self-attention. Reviewing vanilla self-attention formula
in Eq.~\eqref{eq:vanilla}, we found that it uses the full part of
$Q$ and $K$ to calculate a similarity matrix and normalize it.
Kernel-based methods decompose this process as in
Eq.~\eqref{eq:decompose} to approximate $softmax(\cdot)$ such that:
\begin{equation}
    similar(Q, K;\phi) = \phi(Q)\phi(K)^T,
\end{equation}
and we assume that $Q$ and $K$ can continue to be decomposed as a
multiplication of generated feature matrix and themselves:
\begin{equation}
    \phi(Q) = \psi(q)\phi(Q'),
    \label{eq:q decompose}
\end{equation}
\begin{equation}
    \phi(K)^T = \psi(k)\phi(K')^T,
    \label{eq:k decompose}
\end{equation}
where $\psi$ is a kernel function that maps processed data to the
latent low-level feature matrix. After that, the similarity function
can be denoted as secondary decomposed format as:
\begin{equation}
    similar(Q', K', q, k;\phi, \psi) = (\psi(q)\phi(Q')) \cdot
    (\psi(k)\phi(K')^T).
    \label{eq:continue decompose}
\end{equation}
We could merge $\psi(q)$ and $\psi(k)$ via associativity in matrix
product:
\begin{equation}
    similar(Q', K', q, k;\phi, \psi) = \phi(Q') \cdot (\psi(q)\psi(k)) \cdot
    \phi(K')^T.
    \label{eq:merge qk}
\end{equation}
Finally, we denote $\psi(q)\psi(k)$ as the normalization of
low-level similarity matrix $S$ between $Q$ and $K$. Eventually, it
can be seen as
\begin{equation}
    softmax(S) = \psi(q)\psi(k).
\end{equation}
We rearrange the multiplying order to retain the linear complexity.
The ultimate result of self-attention is calculated as:

\begin{equation}
    Attention(Q, K, V, S;\phi) = \phi(Q)((softmax(S)\phi(K)^T)V).
    \label{eq:final self-attention}
\end{equation}

In the vanilla multi-head self-attention, $softmax(\cdot)$ is
applied to normalize each token on the rows of attention matrix.
Compared with Eq.~\eqref{eq:vanilla}, Eq.~\eqref{eq:final
self-attention} shrinks the domain of $softmax(\cdot)$ to low-level
feature similarity matrix $S$. The whole transformation finally
keeps the complexity of self-attention as $O(n)$ as shown in
Figure~\ref{fig:linear attention}. We employ the ELU activation
function as $\phi(\cdot)$ to retain non-negative property as
\cite{DBLP:conf/iclr/QinSDLWLYKZ22} does.

\paragraph{Low-level feature generator}

As aforementioned, attention structure needs to extract representative
low-level feature information to reformulate dot-product
self-attention. In this paper, we propose a low-level feature generator, a CNN-based module to preserve low-level prior knowledge into
a square matrix. As shown in Figure~\ref{fig:network}, We adopt the
basic convolution block as the locality feature extractor and maximum
pooling layer to aggregate the representative feature
from local area. 

Due to practical consideration, we avoid concatenation, splitting
and attention branch during this process, but only apply pooling and
convolution to maintain the module efficiency within self-attention.

\paragraph{Denoising process before self-attention}

We first employ a denoising process on the sequence $X$ after positional embedding, object to approximate original sequence by countering the distribution. Firstly, we sample an expected number of corrupting tokens from a continuous uniform distribution. Specifically, we sample $n$ from $\mathcal{U}([p * l, l])$, where $p$ represent the minimum ratio of the sample number $p \sim \mathcal{U}([0, 1])$ and $l$ denotes the sequence length. Once these values are obtained, we randomly select $n$ pixels to apply the corruption function described below,
\begin{equation}
    X_{c} = (1 - n) * X + n * N, 
\end{equation}
$X$ means original pixel sequence before self-attention and $X_{c}$ indicates which after corruption. Noises $N$ represents a sequence of random values which are generated by discrete uniform distribution $N  = (r_1, r_2, \dots, r_l), r \sim \mathcal{U}({0, 1, \dots, 255})$. 

\section{Experiments}

In this section, we first verify our ESTISR efficiency on real-world applications. Then we conduct experiments on dataset to evaluate our ESTISR performance on recognition accuracy
improvement and image quality. 

\subsection{Dateset}
We select TextZoom~\cite{DBLP:conf/eccv/WangX0WLSB20} as our evaluation
benchmark. TextZoom contains 21740 LR-HR image pairs, which are
collected from two image super-resolution datasets, Real-SR~\cite{DBLP:conf/iccv/CaiZYC019} and SR-RAW~\cite{DBLP:conf/cvpr/ZhangCNK19}. The LR-HR pairs are captured by
cameras with different local length to approximate the real scene.
LR and HR images are resized to $16 \times 64$ and $32 \times 128$,
respectively.

\subsection{Implementation details}
Our ESTISR is implemented on PyTorch. We set mini-batch size to $16$
as each training iteration input with the ADAM optimizer~\cite{DBLP:journals/corr/KingmaB14} for $500$ epochs. The initial
learning rate is set to $4\times10^{-4}$, and decreases half per
$2\times 10^6$ iterations for total $2 \times 10^7$ iterations.
Training is conducted on one NVIDIA Titan X GPU with 2.5GB Memory.
Specifically, we adopt text-focused loss
\cite{DBLP:conf/cvpr/ChenLX21} as loss function to yield better
performance. Text-focused loss is mainly composed of the L2 loss
function with assistance of a pre-trained text recognition module to
highlight position and content.
In order to get precise measurements on efficiency, we gauge the running time of STISR model with $torch.Event$, and record peak memory consumption with
$torch.cuda.max\_memory\_allocated$ that across the STISR-STR process.

\subsection{Comparison with STISR Methods in Efficiency}

\begin{table*}[t!] 
\centering
\begin{tabular}{lcc|cc}
\specialrule{1.5pt}{0pt}{1pt} \multicolumn{1}{c}{Method} & Params (M) & MACs (G) &
Runtime (ms) & Memory (G) \\ \hline TSRN
\cite{DBLP:conf/eccv/WangX0WLSB20} & 2.67 & 0.88    & 39.41 & 1.02
\\ \hline TBSRN \cite{DBLP:conf/cvpr/ChenLX21} & 2.98 & 1.19 & 64.58
& 4.21
\\ \hline TPGSR \cite{DBLP:journals/corr/abs-2106-15368} & 11.88      & 1.72    & 59.72        & 2.68
\\ \hline TATT \cite{DBLP:conf/cvpr/MaLZ22} & 15.94      & 1.99    & 68.49        & 2.15
\\ \hline ESTISR (ours) & 2.16 & 1.21    & \textbf{25.52}        & \textbf{1.02}      \\
\specialrule{1.5pt}{1pt}{0pt}
\end{tabular}
\caption{Efficiency comparison of STISR models.}
\label{tab:efficiency}
\end{table*}

\begin{table*}[htbp] 
\centering \resizebox{2\columnwidth}{!}{
\begin{tabular}{lccccccccccccc}
\specialrule{1.5pt}{0pt}{1pt} \multirow{2}{*}{Method} & \multirow{2}{*}{Loss} &
\multicolumn{4}{c}{ASTER}         & \multicolumn{4}{c}{MORAN}
& \multicolumn{4}{c}{CRNN}          \\ \cline{3-14}
                        &                       & Easy   & Medium & Hard   & Average & Easy   & Medium & Hard   & Average & Easy   & Medium & Hard   & Average \\ \hline
Bicubic                 & -                     & 64.7\% & 42.4\% &
31.2\% & 47.2\%  & 60.6\% & 37.9\% & 30.8\% & 44.1\%  & 36.4\% &
21.1\% & 21.1\% & 26.8\%  \\ \hline SRCNN                   & $L_2$
& 69.4\% & 43.4\% & 32.2\% & 49.5\%  & 63.2\% & 39.0\% & 30.2\% &
45.3\%  & 38.7\% & 21.6\% & 20.9\% & 27.7\%  \\ \hline SRResNet &
$L_2+L_{tv}+L_p$      & 69.4\% & 47.3\% & 34.3\% & 51.3\%  & 60.7\%
& 42.9\% & 32.6\% & 46.3\%  & 39.7\% & 27.6\% & 22.7\% & 30.6\%  \\
\hline TSRN                    & $L_2+L_{GP}$          & 75.1\% &
56.3\% & 40.1\% & 58.3\%  & 70.1\% & 53.3\% & 37.9\% & 54.8\%  &
52.5\% & 38.2\% & 31.4\% & 41.4\%  \\ \hline TBSRN &
$L_{POS}+L_{CON}$     & 75.7\% & 59.9\% & 41.6\% & 60.0\%  & 74.1\%
& 57.0\% & 40.8\% & 58.4\%  & 59.6\% & 47.1\% & 35.3\% & 48.1\%  \\
\hline TPGSR                   & $L_2+L_{TP}$          & 77.0\% &
60.9\% & 42.4\% & 60.9\%  & 72.2\% & 57.8\% & 41.3\% & 57.8\%  &
61.0\% & 49.9\% & 36.7\% & 49.8\%  \\ \hline TATT &
$L_2+L_{TP}+L_{TSC}$  & 78.9\% & 63.4\% & 45.4\% & 63.6\%  & 72.5\%
& 60.2\% & 43.1\% & 59.5\%  & 62.6\% & 53.4\% & 39.8\% & 52.6\%  \\
\hline \hline ESTISR (ours)           & $L_{POS}+L_{CON}$ & 75.8\% &
58.2\% & 42.9\% & 60.0\%  & 71.2\% & 56.2\% & 40.9\% & 57.1\% & 59.7\%  &
47.9\% & 37.2\% & 49.0 \%       \\ \hline \hline HR & -
& 94.2\% & 87.7\% & 76.2\% & 86.6\%  & 91.2\% & 85.3\% & 74.2\% &
84.1\%  & 76.4\% & 75.1\% & 64.6\% & 72.4\%  \\ \specialrule{1.5pt}{1pt}{0pt}
\end{tabular}
}
\caption{Scene text recognition accuracy comparison.}
\label{tab:accuracy}
\end{table*}

\begin{table*}[t]
\centering \resizebox{0.92\linewidth}{!}{
\begin{tabular}{lccccccccc}
\specialrule{1.5pt}{0pt}{1pt} \multirow{2}{*}{Method} & \multirow{2}{*}{Loss} &
\multicolumn{4}{c}{PSNR}        & \multicolumn{4}{c}{SSIM}
\\ \cline{3-10}
                        &                       & Easy  & Medium & Hard  & Average & Easy   & Medium & Hard   & Average \\ \hline
Bicubic                 & -                     & 22.35 & 18.98  &
19.39 & 20.35   & 0.7884 & 0.6254 & 0.6592 & 0.6961  \\ \hline SRCNN
& $L_2$                 & 23.48 & 19.06  & 19.34 & 20.78   & 0.8379
& 0.6323 & 0.6791 & 0.7227  \\ \hline SRResNet                &
$L_2+L_{tv}+L_p$      & 24.36 & 18,88  & 19.29 & 21.03   & 0.8681 &
0.6406 & 0.6911 & 0.7403  \\ \hline TSRN                    &
$L_2+L_{GP}$          & 25.07 & 18.86  & 19.71 & 21.42   & 0.8897 &
0.6676 & 0.7302 & 0.7690  \\ \hline TBSRN                   &
$L_{POS}+L_{CON}$     & 23.46 & 19.17  & 19.68 & 20.91   & 0.8729 &
0.6455 & 0.7452 & 0.7603  \\ \hline TPGSR                   &
$L_2+L_{TP}$          & 23.73 & 18.68  & 20.06 & 20.97   & 0.8805 &
0.6738 & 0.7440 & 0.7719  \\ \hline TATT                    &
$L_2+L_{TP}+L_{TSC}$  & 24.72 & 19.02  & 20.31 & 21.52   & 0.9006 &
0.6911 & 0.7703 & 0.7930  \\ \hline \hline ESTISR (ours) &
$L_{POS}+L_{CON}$   & 23.68 & 19.08  & 19.91 & 21.03   & 0.8836 &
0.6530 &     0.7487 & 0.7678  \\ \specialrule{1.5pt}{1pt}{0pt}
\end{tabular}
} \caption{PSNR and SSIM comparisons.} \label{tab:psnr/ssim}
\end{table*}

We evaluate ESTISR and other STISR methods on the peak memory consumption and running time (speed), which are the most significant efficiency metrics. Others including parameters and MACs reflect the computation and memory occupation but not equal practical results. We select CRNN \cite{DBLP:journals/pami/ShiBY17} as the subsequent STR model. To approximate the model inference under real scene, the full STISR-STR process is conducted
across evaluating dataset of TextZoom
\cite{DBLP:conf/eccv/WangX0WLSB20}. The average running time is calculated by recording each image forwarding time during evaluation.
Results can be seen in Table~\ref{tab:efficiency}, which show
that ESTISR achieves the lowest peak memory consumption and average
running time. Especially, ESTISR reduces 35\% average running time compared with the primal method TSRN, and 60\% of TBSRN, which has the similar network architecture of ESTISR.
Consequences indicate that our RIRB and
softmax shrinking have a positive implication for lightweight STISR
deployment on both resource costs and time-consuming.

\begin{table}[t]
\centering \resizebox{1\columnwidth}{!}{
\begin{tabular}{lcccc}
\specialrule{1.5pt}{0pt}{1pt} Basic block   & Easy    & Medium  & Hard    & Average \\
\hline Conv          & 49.45\% & 40.49\% & 30.81\% & 40.25\% \\
\hline RepVGG        & 49.35\% & 41.01\% & 31.56\% & 40.64\% \\
\hline RepSR block   & 51.64\% & 39.87\% & 31.61\% & 41.04\% \\
\hline ECB           & 52.32\% & 43.50\% & 30.78\% & 42.20\% \\
\hline RIRB-2 (ours) & \textbf{57.55\%} & \textbf{46.09\%} &
\textbf{35.28\%} & \textbf{46.38\%} \\ \hline RIRB-3        &
55.66\% & 45.07\% & 33.25\% & 44.66\% \\ \hline RIRB-4        &
54.31\% & 42.03\% & 32.88\% & 43.07\% \\ \hline RIRB-5        &
52.19\% & 41.75\% & 32.34\% & 42.09\% \\ \specialrule{1.5pt}{1pt}{0pt}
\end{tabular}
} \caption{Recognition accuracy comparisons of structural
re-parameterization blocks. $n$ in RIRB-$n$ denotes expanding
ratio.} \label{tab:re-parameterization}
\end{table}

\subsection{Comparisons with SISR and STISR Methods in Performance}

We compare our ESTISR with current scene text image super-resolution
methods including TSRN \cite{DBLP:conf/eccv/WangX0WLSB20}, TBSRN
\cite{DBLP:conf/cvpr/ChenLX21}, TPGSR
\cite{DBLP:journals/corr/abs-2106-15368}, and TATT
\cite{DBLP:conf/cvpr/MaLZ22}. Besides, we also compare our method
with single image super-resolution methods including SRCNN
\cite{DBLP:journals/pami/DongLHT16} and SRResNet
\cite{DBLP:conf/cvpr/LedigTHCCAATTWS17}.

\textbf{Recognition accuracy improvement.} To measure ESTISR
effectiveness on the improving scene text recognition, we select
three text recognition models, including CRNN
\cite{DBLP:journals/pami/ShiBY17}, ASTER
\cite{DBLP:journals/pami/ShiYWLYB19}, and MORAN
\cite{DBLP:journals/pr/LuoJS19} as the benchmark. As shown in
Table~\ref{tab:accuracy}, ESTISR achieves a competitive
performance on each recognizer. Compared with the baseline TSRN,
ESTISR gets a large improvement in recognition accuracy.
Compared with recent state-of-the-art methods,
ESTISR only sacrifices a little performance degradation, but
achieves a better trade-off between efficiency and performance.

\textbf{Image restoration quality.} We also compare ESTISR with such
methods on the conventional PSNR/SSIM metrics in
Table~\ref{tab:psnr/ssim}. ESTISR has a competitive performance on
SR image results.

\begin{figure*}[t]
    \centering
    \includegraphics[width=1.0\textwidth]{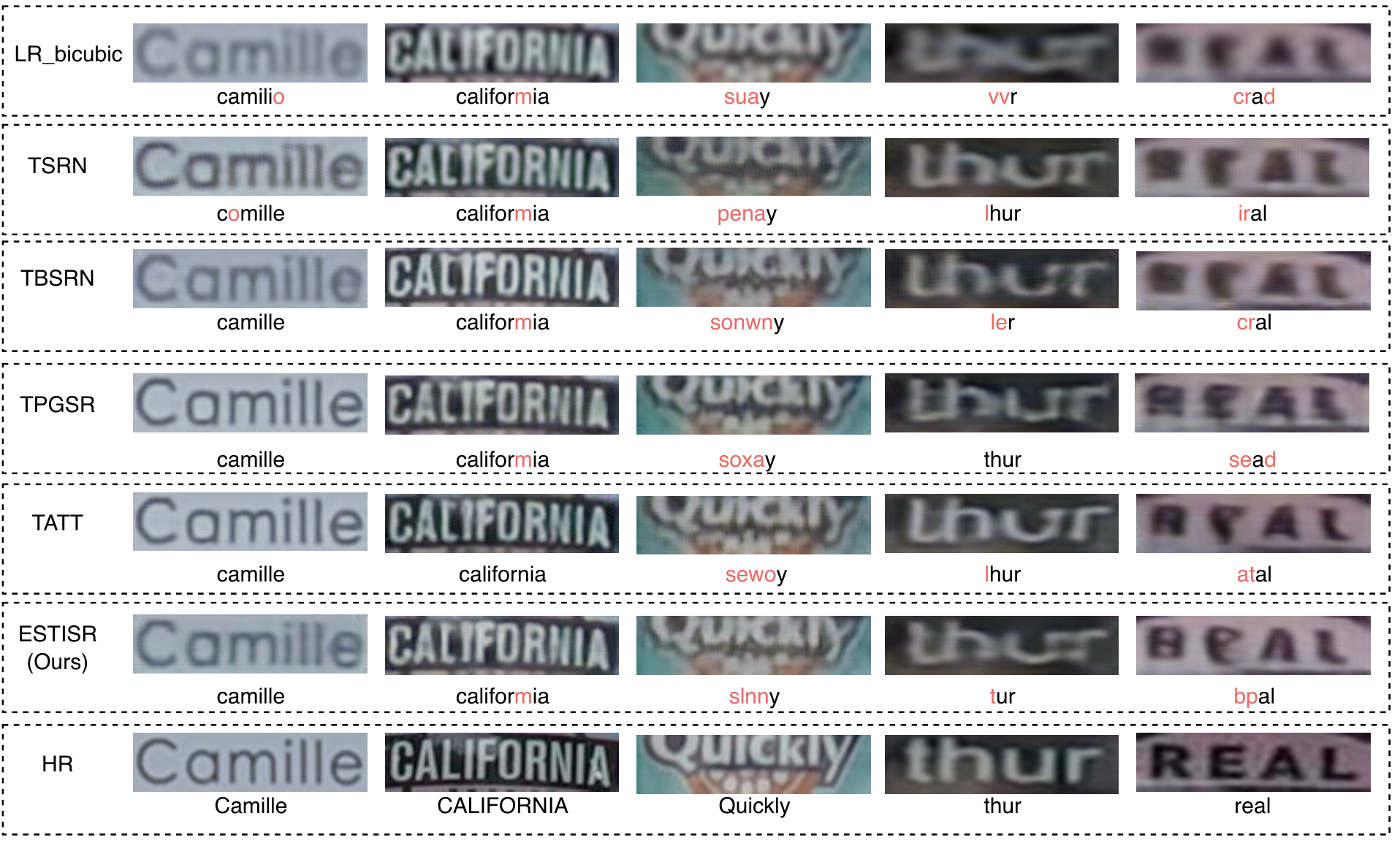}
    \caption{Different STISR models' super-resolution images of samples in Textzoom~\cite{DBLP:conf/eccv/WangX0WLSB20}. LR\_bicubic means low-resolution images after bicubic upsampling. }
    \label{fig:visual}
\end{figure*}

\subsection{Visual comparison}
We visualize several super-resolution (SR) samples of TSRN~\cite{DBLP:conf/eccv/WangX0WLSB20}, TBSRN~\cite{DBLP:conf/cvpr/ChenLX21}, TPGSR~\cite{DBLP:journals/corr/abs-2106-15368}, TATT~\cite{DBLP:conf/cvpr/MaLZ22}, and our ESTISR in Figure~\ref{fig:visual}. Compared with lightweight model TSRN, ESTISR generates higher quality images, and get better accuracy in recognition. Compared with other methods, ESTISR achieves competitive performance in various difficulty. Besides, ESTISR is robust in hard images (see the last column) and incomplete images (see the third column). 

\subsection{Ablation Study}

In this section, we will evaluate the effectiveness of each
component, including the RIRB and softmax shrinking. Experiments are
conducted on TextZoom and recognition accuracy is computed by the
pre-trained CRNN \cite{DBLP:journals/pami/ShiBY17}. All the ablation
experiments are conducted with the same conditions as the above
except for recording the results in 200 epochs.

\subsubsection{Effect of different structural re-parameterization blocks}
We set the standard convolution layer
as the baseline, and then compare
our RIRB with other re-parameterization kernels
including RepVGG \cite{DBLP:conf/cvpr/Ding0MHD021}, RepSR block
\cite{DBLP:conf/mm/WangDS22} and edge-oriented convolution block
(ECB) \cite{DBLP:conf/mm/ZhangZZ21} to explore its effectiveness in
feature extraction. Additionally, we adapt the expanding ratio in
inverted residual from 2 to 6. In Table~\ref{tab:re-parameterization},
we found that RIRB outperforms existing re-parameterization methods.
RIRB has better performance than RepVGG, a re-parameterization block
for high-level tasks. RepSR and ECB perform brilliantly on
SISR as the ConvNet backbone, but for STISR, our RIRB has better
generalization ability at various levels of low-resolution images. According to accuracy changes on expanding ratio, with the expanding ratio
increasing, RIRB is confronted with a performance degradation
gradually. Above all, comparison in Conv and RIRB indicates that structural re-parameterization could largely enhance CNN with no sacrifice of efficiency.

\subsubsection{Effect of softmax shrinking}
In ESTISR and TBSRN \cite{DBLP:conf/cvpr/ChenLX21}, we substitute scales dot-product attention with vanilla
attention to explore the effect of linear self-attention. Particularly, low-level feature generator is removed in
vanilla self-attention. In Table~\ref{tab:linear}, our exxperiments showed that vanilla self-attention is computationally expensive due to its quadratic computation and spatial complexity. Despite this, we found no significant drops in performance. Linear self-attention with softmax shrinking outperformed vanilla self-attention in ESTISR, emphasizing the importance of low-level information in image transformers. Although softmax shrinking requires additional computation, its performance improvement justifies the added $d \times d$ matrix computation cost.

\begin{table}[t]
\centering \resizebox{1\columnwidth}{!}{
\begin{tabular}{llllll}
\specialrule{1.5pt}{0pt}{1pt}
Method  & Runtime (ms) & Memory (G) & Accuracy & PSNR  & SSIM   \\ \hline
TBSRN\_vanilla    & 64.58        & 4.21       & 48.1\%   & 20.91 & 0.7603 \\ \hline
TBSRN\_shrinking  & 43.61        & 1.72       & 45.2\%   & 20.61 & 0.7587 \\ \hline
ESTISR\_vanilla   & 32.36        & 3.52       & 45.3\%   & \textbf{21.02} & \textbf{0.7696} \\ \hline
ESTISR\_shrinking & \textbf{25.52}        & \textbf{1.02}       & \textbf{46.1\%}   & 20.75 & 0.7607 \\ 
\specialrule{1.5pt}{1pt}{0pt}
\end{tabular}
}
\caption{Comparisons between vanilla self-attention and linear self-attention. Accuracy, PSNR and SSIM denote the average value of STISR models which are applied in CRNN \cite{DBLP:journals/pami/ShiBY17}. Suffixal "\_vanilla" and "\_shrinking" represent vanilla self-attention and linear self-attention with softmax shrinking, respectively.}
\label{tab:linear}
\end{table}

\subsection{Different linear self-attention methods.}
We
select efficient self-attention methods with linear complexity as
contrasts, including performer
\cite{DBLP:conf/iclr/ChoromanskiLDSG21}, cosformer
\cite{DBLP:conf/iclr/QinSDLWLYKZ22} and RFA
\cite{DBLP:conf/iclr/Peng0Y0SK21}. 

\textbf{Cosformer \cite{DBLP:conf/iclr/QinSDLWLYKZ22}}. To maintain the non-negative property of attention matrix, cosformer adopts ReLU activation as the representational function $\phi(\cdot)$.
Specifically, we replace positional encoding with cos-based reweighting mechanism as cosformer does. The ultimate similarity has the following form, where $i, j$ represent the position of sequences $Q$ and $K$:
\begin{equation}
    similar(q_i, k_j) = ReLU(q_i)ReLU(k^T_j)cos(\frac{\pi}{2} \times \frac{i - j}{M}),
    \label{eq:cosformer}
\end{equation}
which proposes a novel kernel decomposition relies on non-negative property of $softmax(\cdot)$ attention. 

\textbf{Performer\cite{DBLP:conf/iclr/ChoromanskiLDSG21}}. Performer adopt the fast attention via positive orthogonal random features approach (FAVOR+) algorithm to approximate $softmax(\cdot)$ kernel. Particularly, in experiments we select bidirectional attention mechanism as Transformer backbone, and the same kernel-based method as the softmax shrinking in $softmax(\cdot)$. 

\textbf{Random feature attention (RFA) \cite{DBLP:conf/iclr/Peng0Y0SK21}}. Based on random feature map $\phi(\cdot)$, RFA builds a unbiased estimate to $similar(Q, K)$. Specifically, to overcome the exceed dependency positional encoding, RFA augments history in sequential models with a learned gating mechanism. In general, RFA powers down a part of sequence information via reweighting factors of similarity matrix to improve model complexity.

We conducted a performance comparison of different self-attention formats in our ESTISR model, as summarized in Table~\ref{tab:formers}. Among the evaluated methods, self-attention with softmax shrinking achieved the highest recognition accuracy and performed well on the PSNR/SSIM metrics. Cosformer, which utilizes a similar kernel-based approach, achieved the second best results. However, Performer and RFA exhibited a significant performance degradation, indicating that aggregating sparse information for the $softmax(\cdot)$ kernel is not conducive to low-level tasks. The underlying reasons behind the performance disparities in the low-level field remain unclear. In our future work, we plan to investigate the fundamental principles of efficient transformers in low-level tasks and explore their applicability in diverse domains.

\begin{table}[]
\resizebox{1\columnwidth}{!}{
\begin{tabular}{lccc}
\specialrule{1.5pt}{0pt}{1pt}
Method    & Average accuracy & PSNR  & SSIM   \\ \hline
cosformer & 44.29\%          & 20.61 & 0.7588 \\ \hline
performer & 40.62\%          & 20.13 & 0.7522 \\ \hline
RFA       & 41.09\%          & 20.07 & 0.7412 \\ \hline
softmax shrinking  & \textbf{45.74\%}          & \textbf{20.75} & \textbf{0.7607} \\ 
\specialrule{1.5pt}{1pt}{0pt}
\end{tabular}
}
\caption{Comparisons of different kernel-based methods. Accuracy, PSNR and SSIM denote the average
value of results in STISR models which are applied in CRNN \cite{DBLP:journals/pami/ShiBY17}. }
\label{tab:formers}
\end{table}

\section{Conclusion}

This paper introduces an efficient scene text image super-resolution
(ESTISR) network with the awareness of model efficiency for STISR.
We propose a re-parameterized inverted bottleneck (RIRB) to enhance
the feature extraction module by structural re-parameterization
strategy. Furthermore, we pay attention to the quadratic complexity
of the widely used Transformer in STISR, and propose an softmax shrinking method to
power-down self-attention complexity from $O(n^2)$ to $O(n)$, meanwhile introducing
low-level feature information to improve its
representational capacity. Extensive experiments show that our
ESTISR could largely reduce the time-consuming of STISR, meanwhile
maintaining the image quality and recognition accuracy, achieving a
better trade-off between performance and efficiency.

{\small
\bibliographystyle{ieee_fullname}
\bibliography{paper}
}

\section{Appendix}

\subsection{Pseudo code for RIRB}
We describe how to re-parameterize the re-parameterized inverted residual (RIRB) to standard $3\times3$ convolution by pseudo code. The general process could be divide into three parts: {\bf 1)} Merge Conv1$\times$1-Conv3$\times$3. {\bf 2)} Merge Conv3$\times$3-Conv1$\times$1. {\bf 3)} Merge Conv3$\times$3 and skip connection. Elaboration can be viewed at Algorithm~\ref{alg:RIRB}.

\begin{algorithm}[t] 
    \SetKw{KwFrom}{\ from\ }
    \SetKw{KwFor}{for\ }
    \SetKw{KwReshape}{Reshape\ }
    \SetKw{KwRepeat}{Repeat\ }
    \SetKw{KwPad}{\ pad\ }
    
    \Comment{Merge Conv1$\times$1-Conv3$\times$3}
    \KwReshape $W_0$ \KwFrom $(C_{mid}, C_{in}, 1, 1)$ \KwTo $(C_{in}, C_{mid}, 1, 1)$\;
    $W_{0,1} \gets W_0 \circledast W_1$\;
    \KwReshape $B_0$ \KwFrom $(C_{mid})$ \KwTo $(1, C_{mid}, 1, 1)$\;
    $T \gets ones(1, C_{mid}, 3, 3) \ast B_0$\;
    $B_{0,1} \gets W_1 \circledast T + B_1$\;
    
    \Comment{Merge Conv3$\times$3-Conv1$\times$1}
    \KwReshape $W_{0,1}$ \KwFrom $(C_{mid}, C_{in}, 3, 3)$ \KwTo $(9, C_{mid}, C_{in})$\;
    \KwReshape $W_2$ \KwFrom $(C_{out}, C_{mid}, 1, 1)$ \KwTo $(1, C_{out}, C_{mid})$\;
    \KwRepeat $W_2$ \KwFrom $(1, C_{out}, C_{mid})$ \KwTo $(9, C_{out}, C_{mid})$\;
    $W_{0,1,2} \gets bmm(W_2, W_{0,1})$\;
    \KwReshape $W_{0,1,2}$ \KwFrom $(9, C_{out}, C_{in})$ \KwTo $(C_{out}, C_{in}, 3, 3)$\;

    \Comment{Merge Conv3$\times$3 and skip-connection}
    \eIf{$C_{in}$ == $C_{out}$}
    {
        $W_s \gets zeros(C_{out}, C_{in}, 3, 3)$\;
        $B_s \gets zeros(C_{out})$\;
        \For{$i \ \KwFrom \ 1 \ \KwTo \ C_{out} $}{
            $W_s[0, 0, i, i] \gets 1$\;
        }
    }{
        $W_s \gets W_s \KwPad 0$\;
    }

    $W_{rep} \gets W_{0,1,2} + W_s$\;
    $B_{rep} \gets B_{0,1,2} + B_s$\;

    \caption{Re-parameterization of RIRB. $C_{in}, C_{mid}, C_{out}$ indicate the number of input channels, middle channels, and output channels in RIRB. $\{W, B\}$ denote weight and bias of layer. Indices number indicate the order of covolutional layers in RIRB. $\{W_{s}, B_{s}\}$ corresponds the skip connection of RIRB. $\{W_{rep}, B_{rep}\}$ represents the parameters of re-parameterization result. $zeros$ and $ones$ mean building a new Tensor with $0$ and $1$. $pad$ is the padding operation.}
    \label{alg:RIRB}
\end{algorithm}

\end{document}